\documentclass[pdflatex,sn-mathphys-num]{sn-jnl}

\usepackage{dirtytalk}
\usepackage[T1]{fontenc}
\usepackage[nolist,nohyperlinks,printonlyused]{acronym}
\usepackage{bbold} 
\usepackage{pifont} 
\usepackage{numprint} 
\usepackage{amsmath}
\usepackage{cleveref}
\usepackage{tabularx}
\usepackage{amssymb}
\usepackage[justification=justified]{caption}  
\usepackage{comment}
\usepackage{graphicx}
\usepackage{amsfonts}
\usepackage{lmodern} 
\usepackage[english]{babel}
\usepackage{algorithm}
\usepackage{algorithmicx}
\usepackage{algpseudocode}
\usepackage{mwe}  
\usepackage{xcolor}
\usepackage{mdframed}  
\usepackage{booktabs}

\npthousandsep{\,}

\newcommand{\norm}[1]{\left\lVert #1 \right\rVert}

\newcommand{\mmhead}[1]{\vspace{0.5cm}\noindent\textbf{#1}\vspace{0.3cm}}

\theoremstyle{thmstyleone}%
\theoremstyle{thmstyletwo}%

\theoremstyle{thmstylethree}%

\begin{document}

\title[Ensamble Metamers]{Understanding Cross-Model Perceptual Invariances Through Ensemble Metamers}


\author[1]{\fnm{Lukas} \sur{Lukas} \email{lukas.l.boehm@fau.de}}
\author[1]{\fnm{Jonas Leo} \sur{Mueller} \email{jonas.leo.mueller@fau.de}}
\author[2]{\fnm{Christoffer} \sur{Loeffler} \email{christoffer.loffler@pucv.cl}}
\author[3]{\fnm{Leo} \sur{Schwinn} \email{l.schwinn@tum.de}}
\author[1,4]{\fnm{Bjoern} \sur{Eskofier} \email{bjoern.eskofier@fau.de}}
\author*[1]{\fnm{Dario} \sur{Zanca}}\email{dario.zanca@fau.de}

\affil[1]{\orgdiv{Machine Learning and Data Analytics Lab}, \orgname{Department Artificial Intelligence in Biomedical Engineering, Friedrich-Alexander University}, \orgaddress{\city{Erlangen},  \country{Germany}}}
\affil[2]{\orgdiv{School of Informatics}, \orgname{Pontificia Universidad Católica de Valparaíso}, \orgaddress{\city{Valparaiso}, \country{Chile}}}
\affil[3]{\orgdiv{Data Analytics and Machine Learning Lab}, \orgname{Technische Universität München}, \orgaddress{\city{Munich}, \country{Germany}}}
\affil[4]{\orgdiv{Institute of AI for Health}, \orgname{Helmholtz Zentrum München}, \orgaddress{\city{Munich}, \country{Germany}}}


\abstract{
Understanding the perceptual invariances of artificial neural networks is essential for improving explainability and aligning models with human vision. Metamers—stimuli that are physically distinct yet produce identical neural activations—serve as a valuable tool for investigating these invariances.

We introduce a novel approach to metamer generation by leveraging ensembles of artificial neural networks, capturing shared representational subspaces across diverse architectures, including convolutional neural networks and vision transformers.

To characterize the properties of the generated metamers, we employ a suite of image-based metrics that assess factors such as semantic fidelity and naturalness. Our findings show that convolutional neural networks generate more recognizable and human-like metamers, while vision transformers produce realistic but less transferable metamers, highlighting the impact of architectural biases on representational invariances.
}

\keywords{Metamer Generation, Perceptual Invariances, Artificial Neural Networks}

\maketitle

\section{Introduction}

The computational neurosciences have a long history of analyzing human vision, including its strengths and flaws \cite{marr2010vision}.
Human vision can, for example, be modeled as Bayesian inference over the integration of noisy stimuli \cite{yuille2006vision} into higher level representations.
Computational principles in vision research can be transferred to the computer vision domain, where researchers have spent considerable efforts to understand and interpret the inner workings of visual neural networks in reference to human perception \cite{celeghin2023convolutional, richards2019deep}.
Examples of these efforts include the uncovering of similarities between spatio-temporal cortical dynamics and convolutional neural networks (CNN) \cite{cichy2016comparison}, especially for the lower level processing in the visual system \cite{zheng2018processing}.
Therefore it is reasonable to assume functional similarities in the perceptual invariances between humans and trained CNNs.
While these results offer valuable insights into ecologically rational processing from a computational perspective \cite{todd2012ecological}, deep neural networks exhibit perceptual invariances that differ from human visual processing \cite{feather_model_2023}, particularly in later stages of the visual hierarchy, where processing in the occipital lobe becomes more abstract \cite{zheng2018processing}.

One prominent perceptual phenomenon in vision research used to uncover invariances in humans are \textit{metamers}, pairs of stimuli that are physically distinct, but perceptually indistinguishable \cite{broderick_foveated_2023}.
This phenomenon can help vision researchers understand multiple facets of human vision \cite{rosenholtz2016capabilities}, such as geometric perception of shapes and patterns, color perception through rods and cones, and peripheral vision through the fovea on the retina.
Since human vision and artificial neural network (ANN) models exhibit these functional similarities in learned invariances, a new line of research emerged, where metamers are now studied from a computational perspective on artificial neural networks \cite{feather_metamers_2019, feather_model_2023}.
Studies in this domain are particularly relevant to the ongoing debate about explainability in neural network research \cite{mahendran_understanding_2014,olah_building_2018,olah_feature_2017}, i.e., understanding internal processing of neural networks is the main goal.

Prior work by Feather et al. \cite{feather_model_2023} defined a model metamer as a stimulus that elicits the same activation at some specified layer of a neural network as a reference image but is distinct in its content, thus a different input to the network.
The authors show that the human recognizability of a model’s metamers is strongly predicted by their recognizability by other models, indicating that models possess idiosyncratic invariances beyond those necessary for the task. Building upon these findings we formulate the following claim:

\begin{mdframed}[
  backgroundcolor=gray!10, 
  linecolor=black, 
  roundcorner=5pt, 
  frametitle={}, 
]

Optimizing metamers across multiple models enhances recognizability by prioritizing task-specific invariances while disregarding model-specific idiosyncrasies.

\end{mdframed}

Additionally, Feather et al. \cite{feather_model_2023} highlighted that metamers generated from the later stages of state-of-the-art supervised and unsupervised neural network models were often unrecognizable to humans, raising questions about the alignment of model-based features with human perception, particularly in the context of higher-level, more abstract representations. In this work, we leverage insights from the literature on adversarial attacks \cite{xu2020adversarial} to enhance the optimization scheme used in previous works by employing an optimized Projected Gradient Descent (PGD) method. Additionally,  we expand the analysis to more neural network models, including the vision transformer \cite{khan2022transformers}. Finally, we expand the evaluation strategy proposed by \cite{feather_model_2023} and utilize a comprehensive set of image-based metrics to validate our results. 
While our experiments confirm that metamers from late model layers of single neural networks remain unrecognizable, we expand our understanding of deep learning metamers with the following findings:

\begin{mdframed}[
  backgroundcolor=gray!10, 
  linecolor=black, 
  roundcorner=5pt, 
  frametitle={}, 
]

Multi-model metamers generation using proper optimization schemes can produce naturally looking and recognizable metamers for late model layers of CNN ensembles and robust CNN ensembles.

\end{mdframed}

Counter-intuitively, robust transformers ensembles can generate naturally looking metamers that score low in recognizability, whereas transformers ensembles fail to produce both naturally looking and recognizable metamers, highlighting differences in the way these models represent visual features compared to CNN-based models and highlighting the critical role of architectural biases in shaping representational invariances.

\begin{figure*}
    \centering
    \includegraphics[width=\textwidth]{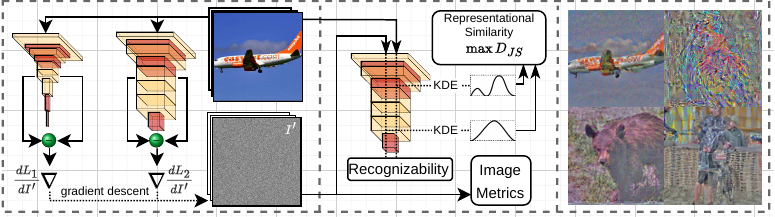}
    \caption{
        (Left) Overview of the metamer generation process. The activations of a model are used to guide the projected gradient descent for a batch of images. The active model is switched every few iterations.
        (Middle) Three techniques are used to evaluate the final metamers. Again, the activations resulting from the reference and metameric stimulus are extracted, turned into a distribution (over the entire dataset/batch), and then compared using the Jensen-Shannon divergence.  Recognizability is an accuracy metric that compares the classification output between reference and metamer. Image Metrics are standalone functions that operate on the final image to rate its visual appearance. They are usually focused on low noise content and natural image elements.
        (Right) Some example metamers generated by a set of CNN models after 5000 steps (late stage).
    }
    \label{fig:figure1}
\end{figure*}

\section{Related Work}

We build upon several areas of prior research relevant to our exploration of model metamers. These include foundational work on model metamers themselves, methods derived from adversarial attack strategies, ensemble learning in neural networks, and a variety of image quality metrics. Below, we review each of these domains to contextualize our contributions.

\mmhead{Model Metamers} 

The seminal work by Feather et al. \cite{feather_metamers_2019, feather_model_2023} investigates model metamers in a single network context.
The core idea is that model metamers are well suited for investigating the divergence of invariances between humans and models.
However, when generating model metamers for a later layer of a deep neural network, the recognizability drops off significantly, indicating that a divergence between models (and humans) arises at later processing stages.
This led to the conclusion that the discrepancy comes from idiosyncratic properties of a specific network, resulting in impaired recognition by both other models and human observers.
These idiosyncrasies are considered to be a result of both the training distribution and the model architecture.
Using robust networks to generate metamers improved the procedure to such an extent that the authors theorized that these idiosyncrasies are completely avoidable \cite{feather_model_2023}.
While the paper used psychophysical trials with human test subjects,
it was also found that other models are just as well suited for evaluating model metamers as humans,
further underlining that idiosyncrasies are the cause for the divergence.

\mmhead{Projected Gradient Descent and Transferability} 

Since adversarial examples were first reported by Szegedy et al. \cite{szegedyIntriguing_2014}, researchers have developed numerous techniques to enhance attack effectiveness~\cite{croce2020reliable, schwinn2021dynamic}.
Notable contributions include Fast Gradient Sign Method (FGSM) \cite{goodfellow_explaining_2014} and Projected Gradient Descent (PGD)~\cite{madry_towards_2019,croce_robustbench_2021}.

We utilize prior work on adversarial attacks for building model metamers, though these concepts have contrasting objectives:
adversarial examples aim to maximize the changes in the output of a network (e.g., cause misclassification) with minimal changes to the input. In contrast, metamers maximize input perturbations while preserving both model outputs and, more importantly, its internal activation patterns.\footnote{
    There is a small semantic difference in the language used for adversarial examples and model metamers.
    While adversarial examples only refer to the perturbation, (model) metamers are defined as a pair of inputs.
    A metamer is a combination of two inputs, in our case, a natural stimulus and a perturbation, that appear the same.
    For clarity, we refer to the perturbation as the metamer and the natural stimulus as the reference.
}

Surprisingly, a large percentage of adversarial examples are transferrable,
meaning they are also misclassified by other models \cite{szegedy_intriguing_2014}.
Transferability is usually improved by using ensemble strategies. 

Previous work on ensemble adversarial attacks has shown that adversarial examples generated by attacking a single model result in poor transferability \cite{chen_adaptive_2023}.
It was also shown that weak models contain invariances that can help strong models in their downstream tasks \cite{chen_explore_2023}, making them viable candidates in any type of ensemble.

\mmhead{Neural network ensembles} 

Research on network ensembles has shown that combining multiple models increases performance in image classification tasks \cite{shen_meal_2019}.
In addition, ensembling has been applied to estimate prediction uncertainties for various machine and deep learning applications \cite{beluch2018power, rigatti2017random}.
The central idea is that models trained on a similar task encode the same functional representation, therefore their predictions are fairly similar, while their discrepancy can give an estimate for prediction confidence.
Although ensemble models do not always have to be from the same architecture, their learned functions must be similar for a successful prediction pipeline.
Therefore, we assume that on a higher level of abstraction, metamers generated from multiple models should span a joint space based on their functional context as the models implement a similar function while their concrete implementation can vary drasticly, e.g., when comparing vision transformers or CNNs \cite{raghu2021vision}.





\mmhead{Image Metrics} 

Image metrics are traditionally used to evaluate the transmission quality of a medium, such as those used to evaluate the quality of generated images, e.g., in the context of generative adversarial networks (GAN) \cite{wang_image_2004,goodfellow_generative_2014}.
Additionally, sophisticated metrics based on neural network outputs have been shown to correlate well with human perception \cite{zhang_unreasonable_2018}.
Therefore, in this paper, we selected a diverse set of nine metrics to evaluate the visual properties of metamers, summarized in \cref{tab:image_metrics}.

\begin{table}[ht]
    \centering
    \begin{tabularx}{\textwidth}{p{4cm} X}
        \toprule
        \textbf{Metric} & \textbf{Description} \\
        \midrule

        Fréchet Inception Distance (FID) & The score is the Fréchet distance between the activations of a target and the reference image, as evaluated by a pre-trained Inception 3 network \cite{heusel_gans_2018}. \\
        
        \midrule
        
        Structural Similarity Index Measure (SSIM) & The measure uses sliding windows to measure image degradation via luminance, contrast, and noise \cite{wang_image_2004}. SSIM was shown to perform poorly on geometric distortions \cite{zhang_unreasonable_2018}. \\
        
        \midrule
        
        Peak Signal-to-Noise Ratio (PSNR) & Expresses the difference in noise between two images \cite{huynh-thu_accuracy_2012}: $\text{PSNR} = 10 \cdot \log_{10} \left( \frac{[\max \text{luminance(x)}]^2}{\text{MSE}(x, y)} \right) \text{dB}$ \\
        
        \midrule
        
        Visual Information Fidelity (VIF) & This calculates differences in a mutual information system as the compared images pass through a human visual system model \cite{sheikh_image_2006}. \\
        
        \midrule
        
        Learned Perceptual Image Patch Similarity (LPIPS) & The measure uses the Euclidean distance to compare the activations between two images to obtain a score that correlates well with human perception \cite{zhang_unreasonable_2018}. \\
        
        \midrule
        
        Relative Average Spectral Error (RASE) & The error metric quantifies the cost of combining two images \cite{wald_data_2002}. \\
        
        \midrule
        
        Spatial Correlation Coefficient (SCC) & Regular images have a low-spectral, but high-spatial resolution \cite{chen_new_2015}. SCC is the Pearson correlation between the variances of the Laplacians \cite{zhou_wavelet_1998}. \\
        
        \midrule
        
        Total Variation (TV) & This is a simple metric for noise calculation that uses the $L_1$ norms of the image gradients \cite{rudin_nonlinear_1992}. \\
        
        \midrule
        
        CLIP Image Quality Assessment (CLIP-IQA) & This is a metric that captures the semantic content of the image using the CLIP model with its contrastively learned text annotations \cite{radford_learning_2021}. Images are evaluated by comparing the reaction to emotional keywords such as \say{natural vs. artificial}, \say{bright vs. dark}, etc. \cite{wang_exploring_2022}. \\
        
        \bottomrule
    \end{tabularx}
    \caption{\textbf{Image Quality Metrics.} This table provides an overview of the nine image quality metrics used in our evaluation. It includes both traditional signal-based approaches and modern deep learning-based techniques. Each metric is briefly described along with its methodological foundation, emphasizing the diversity in how image similarity, fidelity, and perceptual quality are assessed.}
    \label{tab:image_metrics}
\end{table}

\section{Methodology}

\subsection{General Overview}

Neural network metamers are distinct inputs that produce nearly identical activations in a chosen layer of a neural network.
That is, given a reference stimulus $x$, a metameric input $x'$ is an input that elicits activations such that $d(f_l(x), f_l(x'))$ is minimized for some distance metric $d$.
The degree to which $x'$ successfully matches $x$ in terms of activations serves as an indicator of the quality of the neural network metamer \cite{feather_model_2023}.

To generate a metameric input $x'$, we iteratively optimize to minimize the distance between the activations extracted from a selected layer using the reference stimulus and the metameric input. Since this requires direct access to the model activations, it is a white-box scenario.

Let $F_\theta$ be a classification network with weights $\theta$, and let  $f_l$ be the function that extracts the activations of the model $F_\theta$ at a certain layer $l$.
The metamer generation is a non-convex optimization  problem, and we employ the Projected Gradient Descent algorithm to iteratively adjust the metameric stimulus $x'$ at sequential iterations $t$ as follows:
\begin{equation*}
\label{eq:pgd}
    x'_{t+1} = Pr_{x + \mathcal{S}} \left[ x'_t + \alpha \nabla_{x'_t} d(f_l(x'_t), f_l(x)) \right]
\end{equation*}

where $\alpha$ is the step size (or learning rate), and $Pr_{x' + \mathcal{S}}$ is a projection operator that ensures the perturbed input remains within the constrained set $\mathcal{S}$. This set typically is an $\ell_\infty$-norm  projection with a large $\epsilon$ to allow an unrestricted exploration.




\subsection{Multi-Model Metamer Generation}

To address the problem of poor transferability, we propose to extend the original algorithm presented by Feather et al. \cite{feather_metamers_2019} by creating model metamers on multiple models at once.
A simplified outline of the algorithm is described in \cref{alg:multi-model-generation}.

Initially, a reference $x$ from a dataset and randomly initialized synthetic stimuli $x'$ are loaded and passed through a feature-extracting model $F$, where $F \in M$, and $M$ is an ensemble of models.
The two activation maps $F(x)$ and $F(x')$ are then used to calculate a loss,
which can be used to calculate the gradient wrt. the synthetic stimuli $x'$.
The gradient $\nabla$ is then applied to the synthetic stimulus using stochastic gradient descent.
The models are cycled in a round-robin fashion.

\begin{algorithm}
    \begin{algorithmic}[1] 
        \Procedure{MultiModelSolver}{}
        \For{$F \in M$}
        \For{$1 \dots \text{steps}$}
        \State $l \gets L(F(x), F(x'))$
        \State $\nabla \gets \dfrac{\partial l}{\partial x'}$
        \State $x' \gets$ Opt.step($\nabla$)
        \EndFor
        \EndFor
        \State \textbf{return} $l$ \Comment{return the last loss}
        \EndProcedure
        \vspace{1em}
        \Procedure{Metamer Generation}{$x$, $M$, $r$, Scheduler, Initializer}
        \State $x' \gets$ Initializer$(x)$
        \State Scheduler $\gets$ ExponentialDecay($\alpha$, $\gamma$)
        \State Opt $\gets$ Pr(SGD($x'$, Scheduler), $\epsilon=\infty$)
        \For{$1 \dots r$}
        \State loss $\gets$ \Call{Solver}{$x$, $x'$, $M$, Opt, $l$}
        \State Scheduler.step()
        \EndFor
        \State \textbf{return} $x'$  \Comment{Final metamer}
        \EndProcedure
    \end{algorithmic}
    \caption{
        General skeleton for multi-model metamer generation.
    }
    \label{alg:multi-model-generation}
\end{algorithm}


\mmhead{Loss Function}

The loss function is a critical component in the metamer generation process, as it directly influences the gradient calculations.
We optimize relative distances by maximizing the distance between natural and synthetic stimuli, while simultaneously minimizing the difference in their internal representations.
To achieve this, we use the ranking-based loss function Inversion Loss \cite{gomez_exploiting_2020}.

The Inversion Loss aims to minimize the normalized error between activations at a specific stage, defined as ${\norm{a' - a}}/{\norm{a}}$, where $a' = f_l(x')$ is the activation of the metameric input and $a = f_l(x)$ is the activation induced by the reference stimuli at a select stage.
This loss function was also previously used in the foundational work by Feather et al. \cite{feather_model_2023}.


\section{Experiments}

The proposed method for generating model metamers is a generalizable approach that can be applied to any pre-trained classifier. To thoroughly investigate it, we conduct a series of experiments.
First, we demonstrate that the ensemble approach enhances the transferability of model metamers. In the second part, we present evidence that the generated images exhibit significantly superior quality compared to previous methods. Finally, we conduct a comparison between model metamers and targeted adversarial attacks.

\subsection{Experimental Setup}

Throughout our experiments, we selected Stochastic Gradient Descent (SGD), as other optimizers did not yield improved results \cite{feather_model_2023}. Furthermore, an exponential learning rate scheduler was implemented, which progressively reduces the learning rate from an initial value of 1.0 to a minimum of 0.005.

For adversarial attacks, small, imperceptible perturbations are crucial as they are harder to defend against. In contrast, model metamers aim to explore the input space and identify model invariances. Thus, we used a large projection value $\epsilon = 10000$.

\subsection{Datasets}

As most of the models were trained on natural images via ImageNet,
many overlapping invariances between the models lie in this input space of natural images \cite{deng_imagenet_2009}.
For that reason, Geirhos et al. \cite{geirhos_generalisation_2020} compiled a dataset with the intent of being simple and well-recognized by humans.
This simplified dataset with 16 classes \footnote{The 16 classes are: airplane, bicycle, boat, car, chair, dog, knife, truck, bear, bird, bottle, cat, clock, elephant, keyboard}, which are mapped to 231 of the 1000 ImageNet synsets, has been shown to work well in previous psychophysical studies \cite{feather_model_2023}.
Too simplistic datasets like CIFAR-10 might elude many invariances.

\subsection{Model Sets}

Model metamers, like ensemble adversarial attacks, struggle with transferability. Their effectiveness drops significantly when applied to target models with different architectures \cite{chen_adaptive_2023}.
For this reason, we consider different choices for set of models, as described in table \ref{tab:model_sets}. Results for additional model sets are given in Appendix \ref{sec:mixed-model-set}.

VGG19 was chosen as \say{single CNN} as it has proven itself to be the most suitable architecture for gradient-based feature visualization \cite{mordvintsev_differentiable_2018}.
The multimodel sets were composed to feature a diverse set of different sized models.
The set size of 4-5 has been shown to work well in ensemble scenarios \cite{shen_meal_2019}.
In practice, we were also restricted by the availability of robust networks, as we relied on a public repository \cite{croce_robustbench_2021}.
In the appendix \ref{sec:mixed-model-set}, an additional experiment is presented, which attempts to validate the general approach by running cross-validation techniques on many possible model sets. 
All models in use were pre-trained on ImageNet in a supervised fashion and are now set up to take RGB images of size $224 \times 224$.
Feather et al. \cite{feather_model_2023} showed that other training paradigms do not yield significant performance improvements.
The full list of models, including additional ones that were only used for evaluation, can be found in \cref{tab:model-zoo}.

For this evaluation, the model metamers were created using the multi-model approach using the settings described  in \cref{tab:best-settings}.

\begin{table}
    \centering
    \begin{tabular}{|l|l|}
        \hline
        \textbf{Model Set} & \textbf{Models} \\
        \hline
        Single CNN & VGG19 \\
        \hline
        Single Transformer & ViT-B \\
        \hline
        CNN & ConvNext-B, AlexNet, ResNet50, ResNet18 \\
        \hline
        Robust CNNs & ConvNext-B robust, AlexNet Robust, ResNet18 Robust \\
        \hline
        Transformers & Deit-S, ViT-B, Swin-B, XCiT-S \\
        \hline
        Robust Transformers & Deit-S robust, ViT-B robust, Swin-B robust, XCiT-S robust \\
        \hline
    \end{tabular}
    \caption{\textbf{Model sets.} Model sets used for generating ensemble metamers in our experiments.}
    \label{tab:model_sets}
\end{table}

\subsection{Inversion stage}


One of the key parameters in this context is the \emph{inversion stage} $i$, which determines the layer \( l_i \) from which the optimization process begins. 

To balance computational cost while still encuring a sufficiently large dataset for evaluation, we select three intermediate layers from each model. We select layers $i$ corresponding to $20\%$, $50\%$, and $80\%$ of the model's total depth to provide a diverse sampling across the model depth, reducing the likelihood of choosing layers whose representation is not distinct enough from the output logits or the input samples.
Henceforth, we refer to these inversion stages as the \emph{early}, \emph{middle}, and \emph{late} stages, respectively.

\begin{table}
    \centering
    \begin{tabular}{|l|l|}
        \hline
        \textbf{Hyperparameter} & \textbf{Value} \\
        \hline
        Total Steps & 40000 (10 per model, 4000 repetitions) \\
        \hline
        Generation Algorithm & Multi-Model (Round-Robin) \\
        \hline
        Initializer & Uniform $\ell_1$-ball with $\epsilon = 0.1$ (+ Constant $c = 0.5$) \\
        \hline
        LR Scheduler & Exponential (start=1.0, end=0.005) \\
        \hline
        Criterion & Inversion Loss \cite{gomez_exploiting_2020} \\
        \hline
    \end{tabular}
    \caption{\textbf{Hyperparameters.} Settings used to create the best metamers based on prior ablation studies.}
    \label{tab:best-settings}
\end{table}

\subsection{Results}

\begin{figure*}
    \centering
    \includegraphics[width=\textwidth]{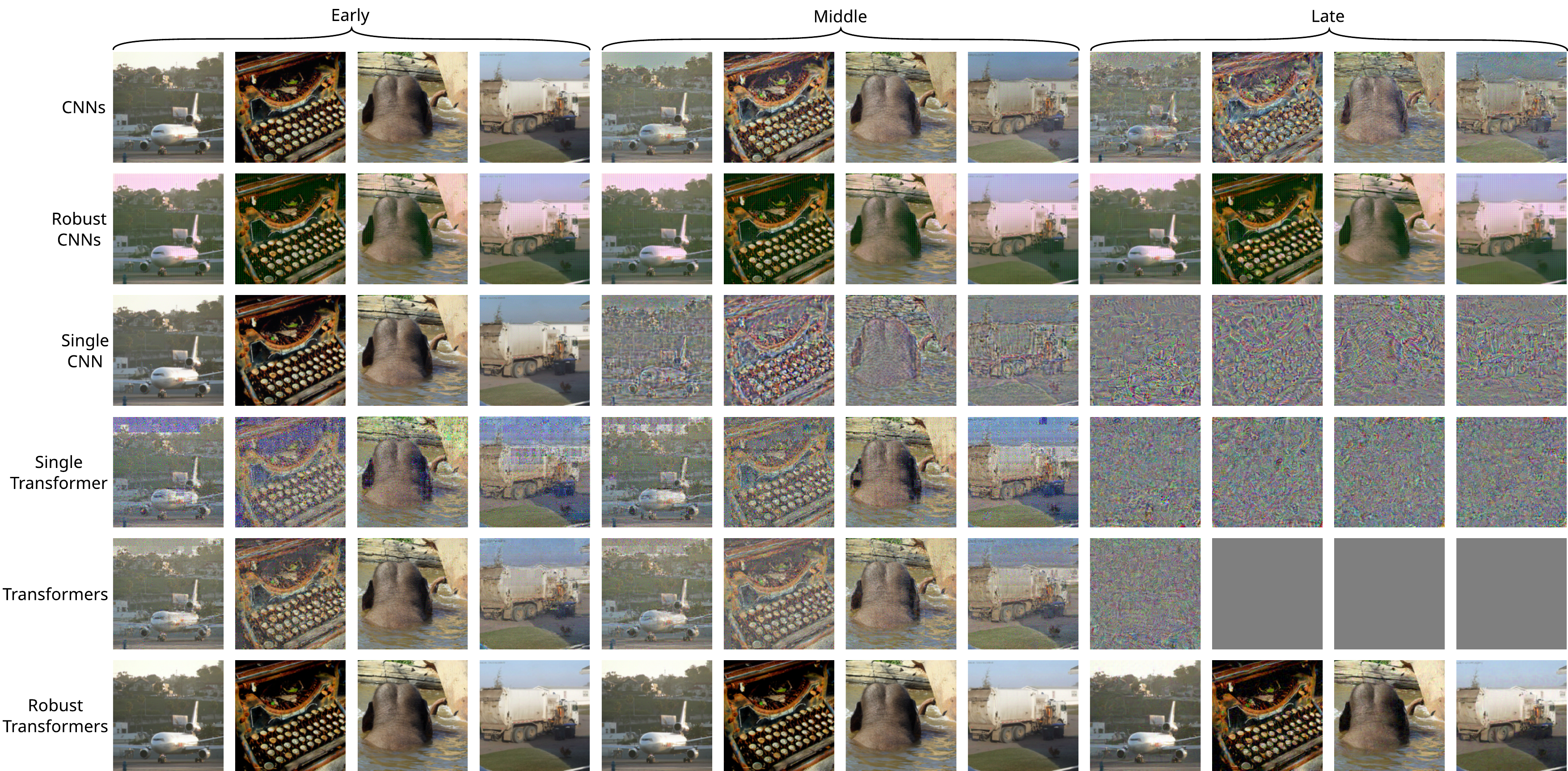}
    \caption{
        \textbf{Examples of single-model and ensemble metamers.}
        Single models tend to produce metamers that quickly become unrecognizable when generated from intermediate or deeper layers. In contrast, ensembles generally yield more robust and recognizable metamers across layers, though this advantage is less pronounced for transformer-based ensembles. Each model or ensemble introduces distinct artifacts.
    }
    \label{fig:best-metamers}
\end{figure*}

Although all model sets successfully generate recognizable metamers in the early stages, a significant divergence is observed in the later stages, as seen in \cref{fig:best-metamers}.
Only the CNN model set and robust ensembles consistently produce recognizable metamers at this stage. 
Notably, the metamers generated by robust ensembles exhibit high visual clarity and a natural appearance, closely resembling precise reconstructions of the reference image, despite not being explicitly optimized for this outcome. 
In contrast, most metamers contain varying levels of noise that degrade visual clarity; however, robust CNNs deviate from this pattern. Rather than exhibiting random noise, these metamers display a repeating checkerboard pattern—most prominent in light regions of the image (e.g., the sky or garbage truck) — which is likely attributable to the effects of overlapping convolutions \cite{odena2016deconvolution}.

\mmhead{Recognizability}

The most common metric to evaluate the transferability of metamers is recognizability.
It is a simple accuracy metric that indicates how well a batch of metamers is recognized by some model.
Specifically, it refers to the percentage of metamers in the batch that had the same output class as the model it was generated on.
We usually evaluate the recognizability using all models in the zoo excluding the generation models, yielding around $25$ accuracy values.

\begin{figure}
    \centering
    \includegraphics[width=\linewidth]{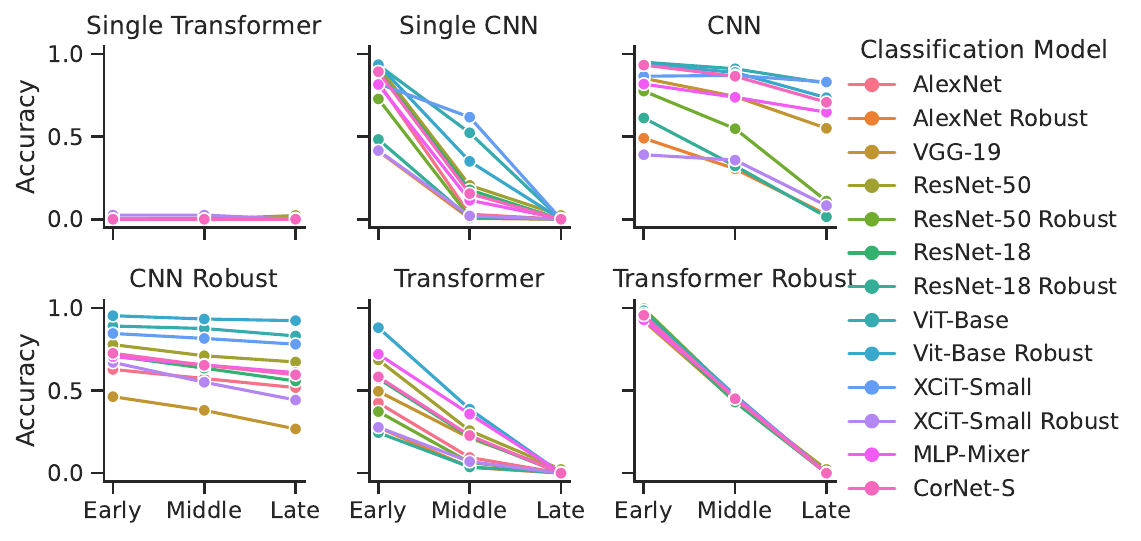}
    \caption{\textbf{Recognizability curves. }Each subplot shows the recognizability curves (accuracy vs. metamer generation stage: Early, Middle, Late) for various model sets. Comparisons include standard and robust variants of CNNs and Transformers. CNN ensembles and robust CNN ensembles retain high recognizability for late stage metamers.}
    \label{fig:recognizability}
\end{figure}

When applied to all parametrizations, we can plot a trend line showing the recognizability across model depth.
When looking at the recognizability plot in \cref{fig:recognizability}, it is visible that the recognizability is retained very well across all stages for model sets that also produced metamers that were recognizable to humans.
This is surprising as the recognizability usually drops off very quickly otherwise.
However, the great recognizability in the late stage comes at the cost of rather low accuracy in the early stage.
When average across all classification models the accuracy is as follows:
$0.477 \pm 0.253$ for early, $0.547 \pm 0.22$ for middle, and $0.419 \pm 0.237$ for the late stage.
In most previous experiments, the recognizability of late-stage metamers was usually below $0.2$.

\mmhead{Image Metrics}

\begin{figure*}
    \centering
    \includegraphics[width=\linewidth]{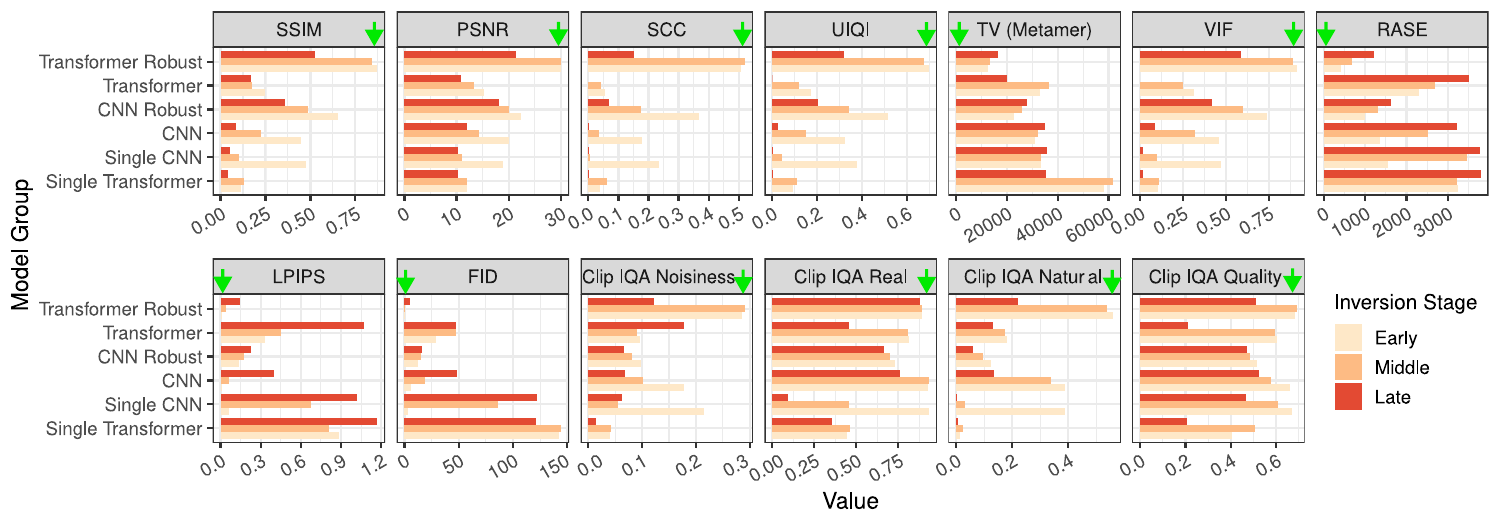}
    \caption{
        \textbf{Image metrics for ensemble metamers.}
        The green arrow indicates whether a larger or lower value indicates better performance. Note that some metrics are normalized resulting in a maximum value of $1.0$. 
    }
    \label{fig:metrics_full}
\end{figure*}

Another way to evaluate model metamers is via image metrics.
They are usually used to evaluate the quality of compression and transmission, as well as generation (e.g. GANs, Diffusion models, etc.).
Here, we use them to evaluate the final metameric image, as seen in \cref{fig:metrics_full}.
Except for FID, LPIPS and the CLIP-IQA score, the metrics do not use a learned classifier to evaluate the image,
yet they are still able to determine the \say{best} metamer.

\mmhead{Representational similarity}

\begin{figure*}[t]
    \centering
    \includegraphics[width=\linewidth]{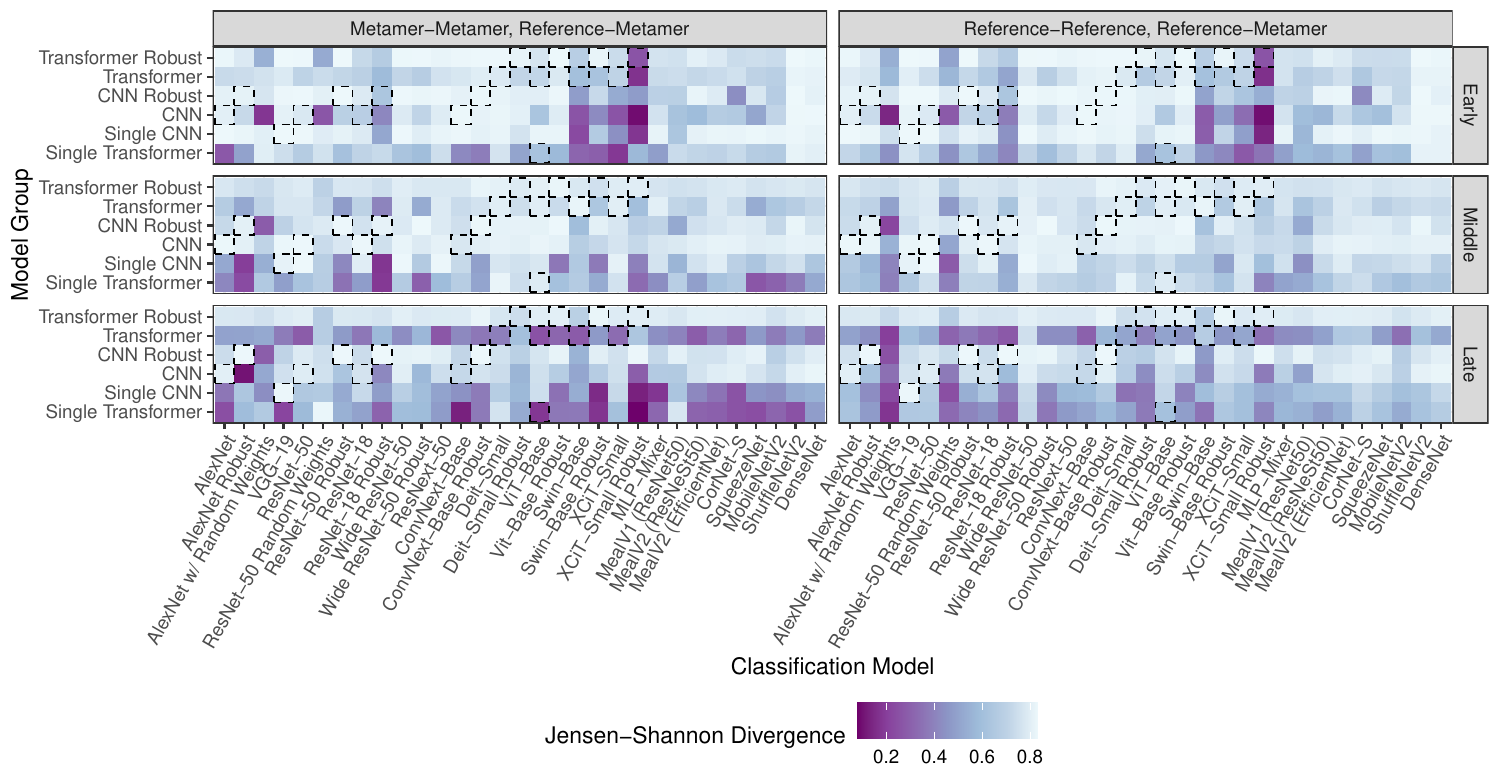}
    \caption{
        \textbf{Jensen-Shannon divergence.}
        The Jensen-Shannon divergence is calculated between each possible pair of representational similarity distributions. Combinations are defined by the model set, the evaluation model, and the generation stage. High divergence values indicate that the distributions have distinct central tendencies. Cells with dashed outlines denote cases where the classification model (row) was included in the corresponding model set (column), rendering the comparison uninformative for our purposes. Such instances do not contribute meaningful insight into metamer transferability.
    }
    \label{fig:repr_sim_heatmap}
\end{figure*}

Recognizability is a primitive score, but works well, as there is a large correlation between the internal activations and the output class.
However, it might not capture all details due to its simplicity.
Rather, we are interested in the activations $a$ and $a'$ that we optimized for above, which represent 
the reference activations and metameric activations.
Across a batch (statistical power test showed >80 is sufficient) we can start to sample from these distributions of activations (${a_k}$ and ${a'_k}$).
We might sample twice from ${a_k}$ to get pairs of reference activations (called \emph{reference-reference}) or do the same with ${a'_k}$ (\emph{metamer-metamer}).
The simplest one is \emph{reference-metamer} which are just the regular (matching) pairs of $a$ and $a'$.

Based on these sampled pairs, we can employ a distance metric (Euclidean distance) to get the representational similarity.
Finally, to be able to reason about them, they are turned into a probability distribution using kernel density estimation (KDE) with a Gaussian kernel and a sampling rate of 1000.
The process is also visualized in \cref{fig:figure1}.

Comparing these probability distributions gives us insights in how the activations of different parametrizations differ. 
Solely comparing the median of the distributions ignores aspects like skewness.
Instead, the Jensen-Shannon divergence is used. \cref{fig:repr_sim_heatmap} shows a heatmap of the divergences.

\subsection{Comparison to Adversarial Examples}


Adversarial examples arise from non-robust, human-imperceptible features in data, and demonstrate their widespread presence in standard datasets~\cite{ilyas_adversarial_2019}.
Recent research shows that adversarial attacks are transferable across different deep learning models, highlighting a fundamental vulnerability in these systems~\cite{madry_towards_2019}.

Adversarial examples also target the desired output (e.g. a certain class) by modifying the input to match a certain representation.
In most cases, adversarial attacks use a limited $\varepsilon$-restricted search space.
Even when confined to only work on noise, the attacks manage to succeed in many cases.

To evaluate whether model metamers behave similar to adversarial examples in a restricted domain,
we created runs based on the best-performing set, the CNN model-mix.
Other settings remain the same, with the addition of a $\varepsilon$ parameter.
We used $8/255 \approx 0.03$ and $0.3$ for, $\varepsilon$ which are both commonly used values in this domain.

\begin{figure*}
    \centering
    \includegraphics[width=\linewidth]{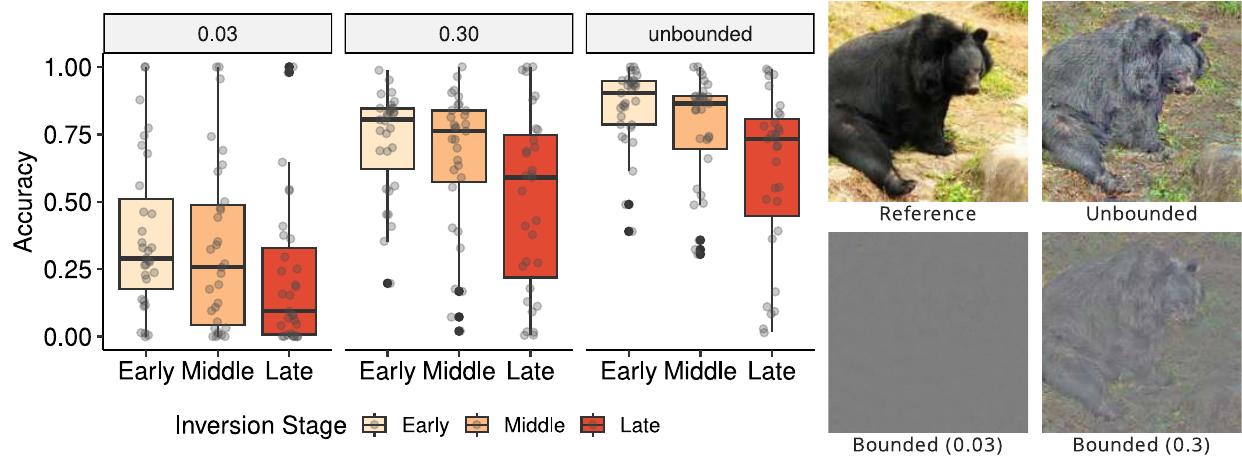}
    \caption{
        (Left) Recognizability values for different $\varepsilon$-values. In total the accuracy was determined by 31 models, including the ones that participated in the generation of the metamers.
        (Right) Example Image across all $\varepsilon$-values generated from the late stage.
    }
    \label{fig:advex_comparison_accuracy}
\end{figure*}

The lowest bound severely limits the effectiveness of the generated metamers (\cref{fig:advex_comparison_accuracy}).
Like a regular adversarial example, the restricted noise causes frequent misclassifications.

\section{Discussion}

\subsection{The Role of Architecture in Metamer Generation}
Our experiments highlight that the choice of architecture plays a critical role in the quality and perceptual validity of generated metamers. CNNs tend to produce metamers that are not only more recognizable but also exhibit a closer resemblance to natural images. This observation can be attributed to the inherent inductive biases of CNNs \cite{wang2023theoretical}, which mimic aspects of early human visual processing—such as local receptive fields and translation invariance—that are essential for capturing the perceptual invariances of natural scenes. In contrast, while vision transformers offer competitive performance on high-level tasks, their lack of spatially localized processing sometimes results in metamers that deviate from human-like perceptual patterns. In fact, CNNs appear to generate metamers that are more “human like,” supporting the idea that architectural design can strongly influence the degree to which artificial representations align with those found in biological vision.

\subsection{Natural Looking Metamers and Their Transferability}
A consistent finding from our analysis is that metamers with a natural appearance tend to transfer best between models. When the metamer retains the inherent statistical properties of natural images, the activations in intermediate layers remain highly compatible across different architectures. This is especially true for standard CNN ensembles, where the shared representation of natural image features results in higher recognizability scores when the metamers are evaluated on models that did not participate in their generation. 
However, robust transformer models yield metamers that are remarkably natural in appearance, closely mirroring the overall texture and statistical properties of real images. This naturalness comes at a cost: the recognizability of these metamers drops when evaluated across different models. One plausible explanation is that the global attention mechanisms in transformers emphasize holistic processing of the image, preserving local textures and naturalistic features \cite{naseer2021intriguing}, while potentially overlooking the fine-grained, discriminative details that other architectures, like CNNs, rely on for robust classification. As a result, although the metamers appear more natural, the subtler cues necessary for accurate recognition are diminished, leading to lower transferability across diverse models.

\subsection{Parallels with Adversarial Attacks}
To draw further conclusions about the aforementioned observations, it is interesting to look at the parallels between the generation of metamers and the construction of adversarial examples. Both techniques leverage gradient-based optimization methods such as PGD to explore the input space, yet they do so with different objectives. Adversarial attacks aim to maximize changes in the network’s output with imperceptibly small perturbations, typically to cause misclassification \cite{chakraborty2021survey}. In contrast, the metamer generation process seeks to identify inputs that produce nearly identical activations—maximizing perceptual invariance—even if this involves larger perturbations. 

Small changes in the image distribution can lead to misclassifications by models used for recognizability, even though the images remain clearly identifiable to human observers (Figure \ref{fig:recognizability}). CNNs, probably due to their similarity to early processing stages of the human visual system, tend to show greater robustness to such perturbations. In general, human visual perception can be conceptualized as the extraction of a semantic hierarchy from the underlying pixel distribution \cite{marr2010vision}, which may contribute to similar robustness in the face of variation through ecologically rational processing \cite{todd2012ecological}. In contrast, vision transformers, which primarily rely on self-attention mechanisms and linear projections, follow a different processing paradigm and may therefore be more susceptible to the accumulation of noise.

This conceptual link between metamer optimization and adversarial attacks is further supported by the effectiveness of ensemble-based methods. Just as ensemble adversarial attacks improve transferability by overcoming architecture-specific vulnerabilities \cite{tramer_ensemble_2020}, generating metamers across multiple models helps mitigate idiosyncratic invariances tied to individual architectures. By encouraging consistency across models, this approach leads to more generalizable metamers and enhances robustness, further illustrating the shared principles underlying both phenomena.

\subsection{Implications for Transferability}
The transferability of metamers is a central metric in our evaluation, directly reflecting the shared representational subspaces among different networks. By employing an ensemble approach, we effectively reduce the influence of model-specific quirks, thereby increasing the likelihood that a metamer generated from one set of models will be recognized correctly by others. Our results indicate that metamers which preserve natural image statistics are more robust in terms of transferability, yielding higher cross-model recognition accuracy. This enhanced transferability not only has implications for understanding the convergence of artificial and human perceptual systems but also opens up practical avenues for improving model interpretability and robustness. Future work should explore how training on diverse and noise-augmented datasets can further align these invariances with those observed in human vision.

\section{Conclusion}

We presented a multimodel metamer generation approach that leverages ensemble methods to improve transferability across deep neural networks. Our results show that CNN-based metamers achieve higher recognizability and align more closely with human visual perception, whereas robust transformer metamers, despite their natural appearance, suffer from reduced recognizability. The primary challenge remains the mitigation of idiosyncratic model invariances, which disrupt cross-model consistency. Addressing these discrepancies—potentially through training on diverse noise types—could further harmonize artificial and human perceptual systems, enhancing both interpretability and robustness.

\section*{Declarations of Conflict of Interest}

The authors declared that they have no conflicts of interest to this work.

\section*{Data availability statement}
The data supporting the findings of this study are derived from publicly available datasets, which are well-known in the research community. These datasets have been properly referenced within the paper. For further information on accessing these datasets, please refer to the references cited within the article.

\section*{Authors contribution statement}
D.Z., J.M., and L.B. conceived the initial idea for the study and designed the experiments. L.B. conducted the experiments, generated the results, and created the plots. D.Z., J.M., C.L., L.S., and B.E. provided supervision. All authors contributed to discussing and revising the experiments and results and writing and reviewing the manuscript. No funding was received to conduct this study.


\bibliography{sn-bibliography}

\begin{appendices}

    \section{Model Zoo}\label{sec:model_zoo}

    \begin{table*}
        \centering
        \begin{tabular}{|l|l|l|}
            \hline
            \textbf{Name}              & \textbf{Dataset}                         & \textbf{Parameters}  \\ \hline \hline
            AlexNet                    & ImageNet                                 & \numprint{61100840}  \\ \hline
            AlexNet Robust             & ImageNet + $L_\infty (\epsilon = 8/255)$ & \numprint{61100840}  \\ \hline
            AlexNet Random             & Randomly Initialized (Kaiming Normal)    & \numprint{61100840}  \\ \hline
            VGG-19                     & ImageNet                                 & \numprint{143667240} \\ \hline
            ResNet-50                  & ImageNet                                 & \numprint{25557032}  \\ \hline
            ResNet-50 Robust           & ImageNet + $L_\infty (\epsilon = 4/255)$ & \numprint{25557032}  \\ \hline
            ResNet-50 Random           & Randomly Initialized (Kaiming Normal)    & \numprint{25557032}  \\ \hline
            ResNet-18                  & ImageNet                                 & \numprint{11689512}  \\ \hline
            ResNet-18 Robust           & ImageNet + $L_\infty (\epsilon = 4/255)$ & \numprint{11689512}  \\ \hline
            ResNext-50                 & ImageNet                                 & \numprint{25028904}  \\ \hline
            ConvNext-Base              & ImageNet                                 & \numprint{88591464}  \\ \hline
            ConvNext-Base Robust       & ImageNet + $L_\infty (\epsilon = 4/255)$ & \numprint{88591464}  \\ \hline
            Deit-Small                 & ImageNet                                 & \numprint{22050664}  \\ \hline
            Deit-Small Robust          & ImageNet + Corruptions                   & \numprint{22050664}  \\ \hline
            ViT-Base                   & ImageNet                                 & \numprint{86567656}  \\ \hline
            Vit-Base (ConvStem) Robust & ImageNet + $L_\infty (\epsilon = 4/255)$ & \numprint{87147112}  \\ \hline
            MLP-Mixer                  & ImageNet                                 & \numprint{59880472}  \\ \hline
            Swin-Base                  & ImageNet                                 & \numprint{87768224}  \\ \hline
            Swin-Base Robust           & ImageNet + $L_\infty (\epsilon = 4/255)$ & \numprint{87768224}  \\ \hline
            XCiT-Small                 & ImageNet                                 & \numprint{26253304}  \\ \hline
            XCiT-Small Robust          & ImageNet + $L_\infty (\epsilon = 4/255)$ & \numprint{26253304}  \\ \hline
            MealV1 (ResNet50)          & ResNet50 + VGG19                         & \numprint{25557032}  \\ \hline
            MealV2 (ResNeSt50)         & SENet154 + ResNet152                     & \numprint{25557032}  \\ \hline
            MealV2 (EfficientNet)      & SENet154 + ResNet152                     & \numprint{5288548}   \\ \hline
            CorNet-S                   & ImageNet                                 & \numprint{53416616}  \\ \hline
            SqueezeNet                 & ImageNet                                 & \numprint{1248424}   \\ \hline
            MobileNetV2                & ImageNet                                 & \numprint{3504872}   \\ \hline
            ShuffleNetV2               & ImageNet                                 & \numprint{2278604}   \\ \hline
            DenseNet                   & ImageNet                                 & \numprint{7978856}   \\ \hline
        \end{tabular}
        \caption[The model-zoo. Not all of them are viable for generating metamers.]{
            The model-zoo. Not all of them are viable for generating metamers.
            All adversarially trained models have \say{Robust} appended to their name.
            The robust models have been trained on ImageNet with additional perturbations (as specified in the table).
            \say{Corruptions} refers to a new training paradigm where the model is trained on a dataset with additional perturbations instead of perturbations in the normed $\epsilon$-ball.
            The corruption datasets are \say{ImageNet-C} and \say{ImageNet-3DCC} \cite{croce_robustbench_2021}.
            In the case of the MEAL-based ensemble models, we mention the teacher models instead of the dataset.
        }
        \label{tab:model-zoo}
    \end{table*}


    Contrary to other studies, it is not possible to restrict ourselves to one representative model per architecture type.
    A model zoo of 31 models from various architectures was compiled.
    The models are listed in \cref{tab:model-zoo}.
    We explicitly only included models that were trained in a supervised manner,
    as Feather et al. \cite{feather_model_2023} found that there is no significant impact of training style on metamer generation.
    Models were sourced from \emph{timm} \cite{wightman_pytorch_2019} and PyTorch Hub \cite{paszke_pytorch_2019}.
    Robust models are downloaded from the RobustBench repository, a large collection of adversarially trained models \cite{croce_robustbench_2021}.
    The only exception is AlexNet Robust, which is included from \cite{feather_model_2023}, hence the different training parameters.
    We attempted to find a robust counterpart for each model.
    However, due to time and resource constraints, we had to rely on publicly available models and were not able to train our own adversarially robust models.

    \section{Mixed-Model}\label{sec:mixed-model-set}

    The rather large model zoo allows for a large amount of model combinations.
    Just picking 5 of them yields an intractable amount ${{31}\choose{5}} = 169.911$.
    The experiment above also did not mix robust and non-robust models.
    Inspired by Monte Carlo cross-validation, we generated 100 randomly sampled model sets with sizes varying between 2 and 10 models.
    Each model set is executed for $10000$ steps at the middle stage, as we assume there is sufficient convergence at this point.
    To speed up the computation time, the dataset was further reduced to 80 images.

\begin{figure}
        \centering
        \includegraphics[width=1\linewidth]{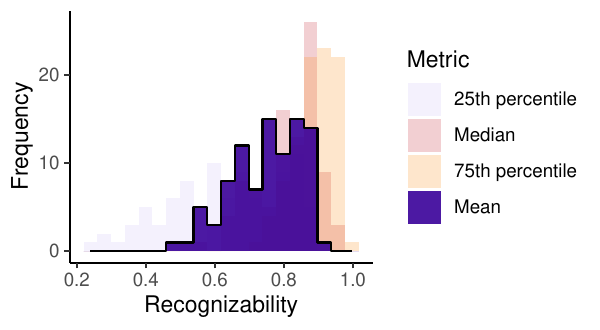}
        \caption{Accuracy distributions for 100 different model combinations. Metamers were rated by all models in the model zoo, except for the ones that generated them. This means that there isn't a fixed set of evaluation models.}
        \label{fig:mixed_model_sets_accuracy}
    \end{figure}

    \cref{fig:mixed_model_sets_accuracy} shows that mixed model sets still boast an impressive recognizability, on average $0.751 \pm 0.101$.

    Besides showing the generalizability of this approach, we also aim to find out whether there are certain models that increase/decrease the performance of each model set they are part of.
    To this end we employed a simple Difference-in-means approach.
    For each model $M$ an apportionment based on the presence of $M$ in the runs is compiled.
    The averaged difference in recognizability between the present and absent set gives us the difference in means $\Delta_M \mathbb{E}$. The data is displayed in \cref{tab:mixed_model_sets}.
    As argued above, robust models have a positive impact while transformers are conspicuous due to their negative performance impact

    \begin{table}
        \begin{tabular}{lll}
        \multicolumn{3}{l}{Model \hspace{1.9cm} Appearances \quad Difference In Means} \\ \hline
        Swin-Base Robust          & 26          & 0.118               \\
        Vit-Base Robust           & 19          & 0.089               \\
        Wide ResNet-50 Robust     & 17          & 0.083               \\
        ResNet-18 Robust          & 24          & 0.049               \\
        MLP-Mixer                 & 19          & 0.048               \\
        ViT-Base                  & 16          & 0.039               \\
        ResNet-50 Robust          & 21          & 0.038               \\
        MealV2 (ResNeSt50)        & 21          & 0.036               \\
        XCiT-Small Robust         & 17          & 0.036               \\
        ConvNext-Base Robust      & 15          & 0.035               \\
        Deit-Small Robust         & 21          & 0.033               \\
        VGG-19                    & 16          & 0.033               \\
        MealV2 (EfficientNet)     & 17          & 0.026               \\
        Deit-Small                & 21          & 0.025               \\
        AlexNet                   & 21          & 0.022               \\
        MobileNetV2               & 19          & 0.015               \\
        ResNext-50                & 18          & 0.014               \\
        ConvNext-Base             & 19          & 0.011               \\
        SqueezeNet                & 16          & 0.011               \\
        Wide ResNet-50            & 21          & 0.010               \\
        AlexNet w/ Random Weights & 10          & 0.009               \\
        CorNet-S                  & 21          & 0.009               \\
        DenseNet                  & 17          & 0.008               \\
        ResNet-50                 & 16          & 0.008               \\
        AlexNet Robust            & 17          & 0.005               \\
        ShuffleNetV2              & 13          & 0.004               \\
        XCiT-Small                & 19          & 0.004               \\
        ResNet-50 Random Weights  & 15          & 0.003               \\
        Swin-Base                 & 16          & 0.002               \\
        ResNet-18                 & 13          & 0.001               \\
        MealV1 (ResNet50)         & 21          & 0.000
        \end{tabular}
        \label{tab:mixed_model_sets}
        \caption{
        The presence of all models across the 100 validation runs.
        To determine the impact of each model on the generation results, a simple Difference in Means approach using the recognizability data was used.
        }
    \end{table}

\end{appendices}

\end{document}